\documentclass[conference]{IEEEtran}
\IEEEoverridecommandlockouts
\usepackage{cite}
\usepackage{siunitx}
\usepackage{amsmath,amssymb,amsfonts}
\usepackage{algorithm}
\usepackage{algorithmic}
\usepackage{graphicx}
\usepackage{textcomp}
\usepackage{xcolor}
\usepackage{array}
\usepackage{balance}
\usepackage{booktabs}

\usepackage{soul}

\newcommand{\vect}[1]{\boldsymbol{#1}}
\newcommand{\mat}[1]{\mathbf{#1}}

\newcommand{\figplaceholder}[2]{%
  \IfFileExists{#1}{\includegraphics[width=\columnwidth]{#1}}{%
    \fbox{\parbox[c][1.78in][c]{0.93\columnwidth}{\centering\footnotesize #2}}%
  }%
}
\newcommand{\widefigplaceholder}[2]{%
  \IfFileExists{#1}{\includegraphics[width=\textwidth]{#1}}{%
    \fbox{\parbox[c][2.08in][c]{0.96\textwidth}{\centering\footnotesize #2}}%
  }%
}

\def\BibTeX{{\rm B\kern-.05em{\sc i\kern-.025em b}\kern-.08em
    T\kern-.1667em\lower.7ex\hbox{E}\kern-.125emX}}
\begin{document}

\title{Structure-Actuation Co-Design of All-Electric Heavy-Duty Robotic Manipulators\\

}

\author{\IEEEauthorblockN{Mohammad Bahari}
\IEEEauthorblockA{\textit{Engineering and Natural Sciences} \\
\textit{Tampere University}\\
Tampere, Finland \\
mohammad.bahari@tuni.fi}
\and
\IEEEauthorblockN{Amir Hossein Barjini}
\IEEEauthorblockA{\textit{Engineering and Natural Sciences} \\
\textit{Tampere University}\\
Tampere, Finland \\
amirhossein.barjini@tuni.fi}
\and
\IEEEauthorblockN{Jouni Mattila}
\IEEEauthorblockA{\textit{Engineering and Natural Sciences} \\
\textit{Tampere University}\\
Tampere, Finland \\
jouni.mattila@tuni.fi}
}

\maketitle

\begin{abstract}
This paper presents a mixed discrete--continuous multiobjective
bilevel framework for the structure--actuation co-design of
all-electric heavy-duty robotic manipulators. The upper level
optimizes manipulator geometry and actuation topology, while
joint-specific lower-level searches construct feasible archives
of discrete motor, gearbox, and screw configurations. A
system-level coordinator selects actuator packages according to
hardware mass, co-design cost, and loss-aware supply energy, and
the resulting actuator masses are fed back to the robot model
until a mass-consistent fixed point is obtained. Evaluation on a
full-pose $SE(3)$ benchmark with a 100-kg tool payload yields
eight feasible nondominated designs, all satisfying the 12
prescribed poses. Actuator-mass coupling converges in three iterations
for every reported design. A fixed-design ablation shows that
one-pass sequential sizing changes hardware mass by up to
2.13\%, co-design cost by 10.20\%, and supply energy by 9.40\%
in magnitude. Payload screening further reveals increasing
mass--cost--energy demand and changes in competitive mechanism
topology with carried load.
\end{abstract}


\section{Introduction}

Heavy-duty robotic manipulators require high payload capacity,
extended reach, and reliable operation under demanding loads,
for which hydraulic actuation has traditionally been attractive
\cite{petrovic2022mathematical}. Advances in electromechanical
actuators (EMAs), together with the efficiency limitations of
hydraulic systems, are motivating all-electric alternatives
\cite{li2021study,bahari2026electrification,bahari2025all}. Electrification,
however, is not simply an actuator-replacement problem.
Manipulator geometry and mechanism topology determine
workspace, mechanical advantage, structural loading, and
actuator demand \cite{wu2025exact,nabavi2020parametric,wu2024transmission}, whereas the
selected motor--transmission packages determine actuator mass,
reflected inertia, torque--speed capability, and efficiency
\cite{park2023selection,10199841}. Since these package masses
subsequently modify the robot dynamics and structural loading,
mechanical design and actuator selection form a coupled
structure--actuation problem rather than sequential sizing
stages \cite{lu2025differentiable}.

Integrated robot design has consequently received increasing
attention. Simulation-aided and cascaded formulations have
jointly optimized hardware and motion variables
\cite{fadini2022simulation,peralta2024design,barjini2025surrogate}, while bilevel
optimization has been applied to morphology--behavior
coordination \cite{kim2021mo}. System-level electrical-drive
optimization has likewise demonstrated the value of coordinating
interacting component designs \cite{lei2014system}. Discrete
component selection \cite{hoffman2025addressing}, joint
motor--motion optimization \cite{stein2025application},
linkage-energy optimization \cite{paz2024energy}, and bilevel
coordination of manipulator operation and EMLA efficiency
\cite{11008622} have further extended this direction. 

This paper develops a mixed discrete--continuous multiobjective
bilevel framework for structure--actuation co-design. The upper
level optimizes an 11-variable mechanical design comprising
link dimensions, mechanism topology, and linkage geometry,
while joint-specific lower-level searches construct bounded
archives of feasible EMA/EMLA packages. Their Cartesian product
is coordinated at the system level according to hardware mass,
co-design cost, and loss-aware supply energy. The selected
actuator masses are then fed back into the rigid-body and
structural models until an actuator-mass fixed point is reached.
Evaluation uses a full-pose $SE(3)$ benchmark, a nominal
co-design study, payload-sensitivity, and a
fixed-design ablation of the mass-feedback loop.

The remainder of this paper is organized as follows. Section~II presents the system model and benchmark, Section~III describes the proposed bilevel co-design framework, Section~IV discusses the results, and Section~V concludes the paper.

\section{System Model and Benchmark}

\subsection{Architecture and Task}

The manipulator comprises a three-DoF co-designed main arm and
a fixed-design three-DoF $X$--$Y$--$X$ spherical wrist. The
main-arm coordinates are
$\vect q=[q_1,q_2,q_3]^\mathsf{T}$; joint~1 is rotary, whereas
joints~2 and~3 may employ direct rotary actuation or three-bar or four-bar mechanisms. The wrist provides end-effector orientation but is
excluded from the co-design variables.

\begin{figure}[t]
    \centering
    \includegraphics[width=0.75\columnwidth]
    {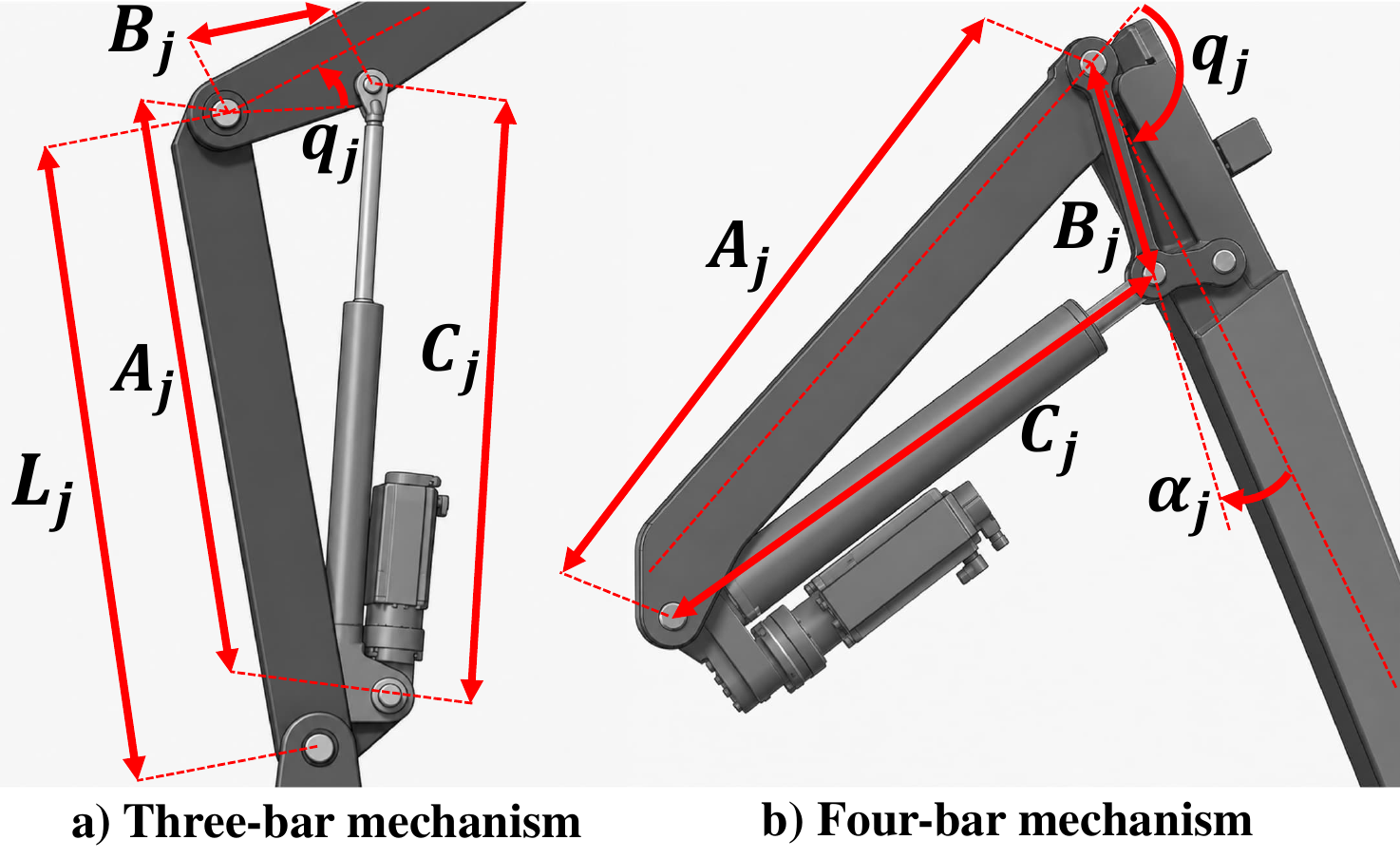}
    \caption{Geometric parameterization of the candidate
    linkage mechanisms: (a) three-bar and (b) four-bar
    mechanisms.}
    \label{fig:joint_mechanisms}
\end{figure}

The upper-level mechanical design is
\begin{equation}
\vect{x}=
[L_1,L_2,L_3,\nu_2,\nu_3,
A_2,B_2,\alpha_2,A_3,B_3,\alpha_3]^\mathsf{T},
\label{eq:x}
\end{equation}
where $\nu_j\in\{0,1,2\}$ denotes rotary, three-bar, and
four-bar mechanisms, respectively. For linearly actuated joints,
$A_j$ and $B_j$ define the linkage geometry and $\alpha_j$
provides the additional four-bar offset. Inactive variables are
assigned canonical values before evaluation. The candidate joint mechanisms and their geometric
parameterization are illustrated in
Fig.~\ref{fig:joint_mechanisms}.

The benchmark specifies 12 full TCP poses,
\begin{equation}
\mat T_{d,i}=
\begin{bmatrix}
\mat R_{d,i} & \vect p_{d,i}\\
\vect 0^\mathsf{T} & 1
\end{bmatrix},
\qquad i=1,\ldots,12,
\label{eq:pose}
\end{equation}
for which the complete arm--wrist chain is checked for
kinematic feasibility, including joint limits, wrist
singularity, solution continuity, and main-arm conditioning.
For dynamic actuator sizing, the 12 TCP positions are closed
into a 13-waypoint cycle and tracked using damped closed-loop
inverse kinematics \cite{yu2024inverse}. The wrist remains at its nominal configuration during this cycle; hence its actuator dynamics
and energy are excluded. Tracking feasibility is quantified by
\begin{equation}
e_{\max}=\max_k
\left\|
\vect p_d[k]-\vect p_e[k]
\right\|_2 .
\label{eq:tracking}
\end{equation}

\begin{figure}[t]
    \centering
    \includegraphics[width=0.95\columnwidth]
    {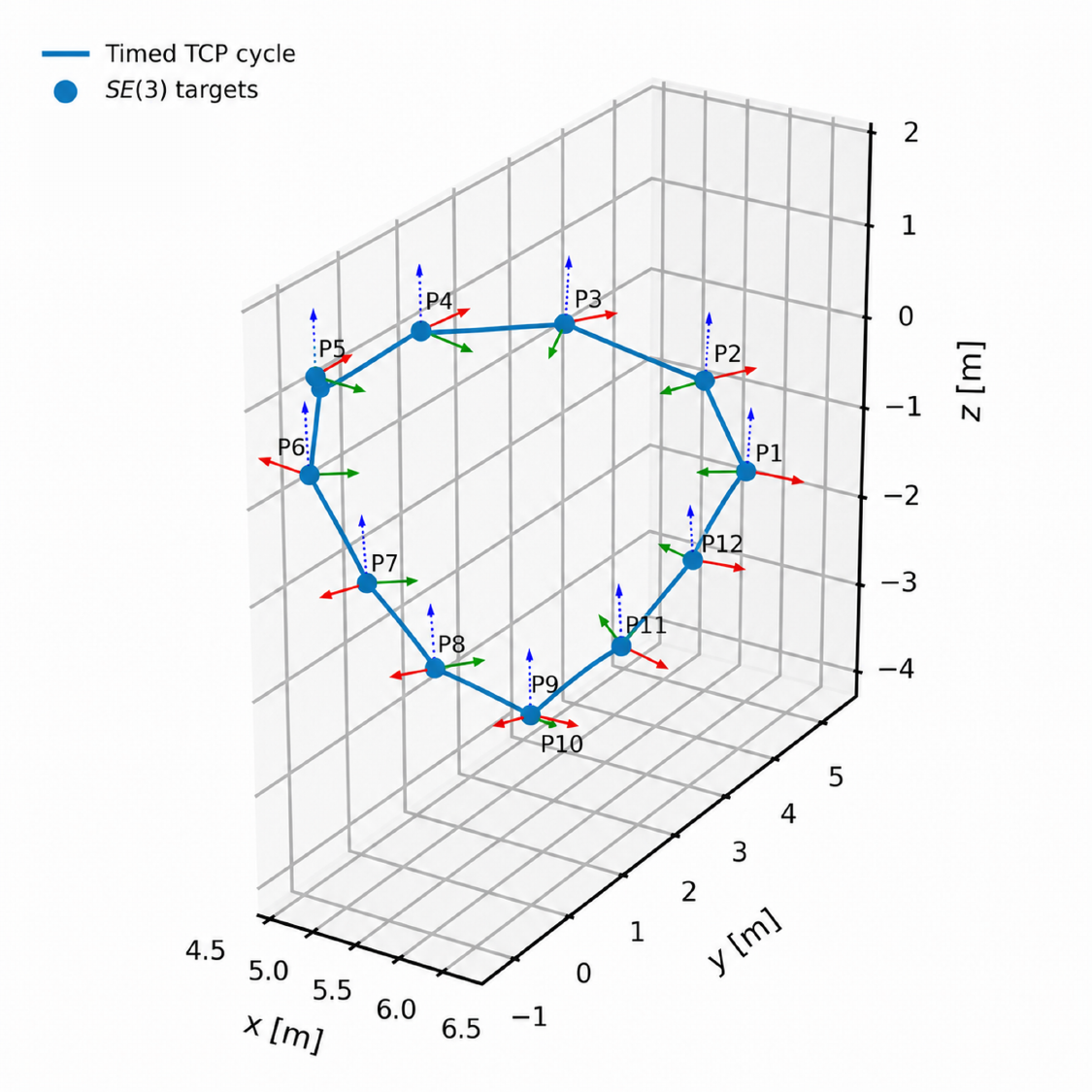}
    \caption{Full-pose $SE(3)$ benchmark with 12 prescribed TCP
    positions and orientations and the closed dynamic-sizing
    trajectory.}
    \label{fig:benchmark}
\end{figure}

\subsection{Dynamics, Linkages, and Structural Screen}

For a repaired design $\vect x$ and actuator-mass vector
$\vect m_a$, the main-arm dynamics are
\begin{equation}
\vect\tau=
\mat M(\vect q;\vect x,\vect m_a,m_d)\ddot{\vect q}
+\vect h(\vect q,\dot{\vect q};\vect x,\vect m_a,m_d),
\label{eq:dynamics}
\end{equation}
where $m_d=m_{\rm wrist}+m_{\rm tool}$ is the distal carried
mass. The wrist contributes to hardware mass whereas the tool
does not; both contribute to dynamic and structural loading.

For an EMLA-driven joint, the effective linkage angle is
\begin{equation}
q_{e,j}=
\begin{cases}
\delta_j+q_j, & \nu_j=1,\\
\delta_j+q_j-\alpha_j, & \nu_j=2,
\end{cases}
\quad
\delta_2=\frac{\pi}{2},\ \delta_3=\pi ,
\label{eq:effective_angle}
\end{equation}
and the geometry, velocity, and regularized force estimate
are given below, with $\epsilon_C$ denoting a numerical
regularization parameter in metres:
\begin{equation}
\begin{aligned}
C_j&=\sqrt{A_j^2+B_j^2-2A_jB_j\cos q_{e,j}},\\
g_j(q_j)&=\frac{A_jB_j\sin q_{e,j}}{C_j},\\
v_{a,j}&=g_j(q_j)\dot q_j,\qquad
F_{a,j}=\frac{\tau_j}{g_j(q_j)+\epsilon_C}.
\end{aligned}
\label{eq:linkage_transmission}
\end{equation}
Invalid installation geometries or insufficient actuator-length
margins are rejected before dynamic evaluation.

All links use a fixed hollow-square steel section with
$b_o=0.150$~m, $t=0.005$~m,
$b_i=b_o-2t=0.140$~m, and
$\rho=7850$~kg/m$^3$. Its properties are
\begin{equation}
A_s=b_o^2-b_i^2,\qquad
I_s=\frac{b_o^4-b_i^4}{12},\qquad
m_{L,i}=\rho A_sL_i .
\label{eq:section_properties}
\end{equation}
The section is fixed rather than optimized. Structural
feasibility is screened using
\begin{equation}
\sigma_{\max}=
\max_{i,k}
\left(
\frac{|F_{\mathrm{ax},i}[k]|}{A_s}
+\frac{M_{b,i}[k](b_o/2)}{I_s}
\right),
\label{eq:stress}
\end{equation}
including downstream link, actuator, wrist, and tool masses.
This axial-plus-bending model is an engineering feasibility
screen rather than a local-stress, fatigue, or buckling
certification model.

\subsection{Discrete Actuator Packages and Loss-Aware Energy}

For joint $j$, the lower-level decisions are
$\vect y_j^{\rm EMA}=[i_m,i_{N_g}]^\mathsf{T}$ and
$\vect y_j^{\rm EMLA}=[i_m,i_{N_g},i_p]^\mathsf{T}$,
selecting a catalogue motor, gearbox ratio $N_g$, and, for an
EMLA, screw lead $p$. With $N_g$ denoting the motor-to-output speed ratio,
$\omega_{m,j}=N_g\dot q_j$ for EMAs and
$\omega_{m,j}=(2\pi N_g/p)v_{a,j}$ for EMLAs,
where $p$ is expressed in m/rev. Complete
variable package mass and cost are
\begin{equation}
\begin{aligned}
m_{a,j}&=m_m+m_g+\mathbb{I}_{\rm EMLA}m_s+m_{\rm aux},\\
C_{a,j}&=C_m+C_g+\mathbb{I}_{\rm EMLA}C_s+C_{\rm aux},
\end{aligned}
\label{eq:package_mass_cost}
\end{equation}
including auxiliary hardware. Motor torque, speed, power, and current limits are enforced,
while EMLA screw feasibility also includes axial strength,
buckling, and critical speed \cite{jang2025selection}.

Supply energy is estimated using a catalogue-calibrated
load-dependent loss surrogate,
\begin{equation}
\begin{aligned}
P_{\rm loss,N}&=P_N(\eta_N^{-1}-1),\\
P_{\rm loss}(t)&=P_{\rm loss,N}
\left(\frac{\tau_m(t)}{\tau_N}\right)^2,\\
E_{\rm sup}&=\int
\max\!\left[P_m(t)+P_{\rm loss}(t),0\right]dt ,
\end{aligned}
\label{eq:energy}
\end{equation}
where $P_m=\tau_m\omega_m$ is signed mechanical motor power.
Negative net power is not credited as regeneration. The model
is intended for comparative actuator selection; speed-dependent
iron, windage, inverter, and thermal losses are not explicitly
resolved. Wrist-actuator energy is likewise excluded because
the timed dynamic model considers only the co-designed main arm.

\section{Mass-Consistent Bilevel Co-Design}

The mechanical design is the leader decision, while actuator
selection provides its mass-consistent follower response:
\begin{equation}
\begin{aligned}
\underset{\vect x\in\mathcal X}{\operatorname{minimize}}
\quad&
\vect F_u\!\left(\vect x,\vect y_b^\star(\vect x)\right)\\
\text{subject to}\quad&
\vect G_u\!\left(\vect x,\vect y_b^\star(\vect x)\right)
\preceq \vect 0 ,
\end{aligned}
\label{eq:bilevel}
\end{equation}
where $\vect y_b^\star$ is the converged three-joint actuator
package obtained through joint-level search, system-level
coordination, and actuator-mass feedback. Hence, each leader
candidate is evaluated under the mass distribution induced by
its selected actuator packages.

\subsection{Joint-Level Actuator Search}

For given $(\vect x,\vect m_a)$, the robot simulation provides
the joint histories used for actuator selection. The
family-dependent lower objectives are
\begin{equation}
\vect F_{\ell,j}=
\begin{cases}
[P_N^{\rm kW},\,N_g,\,\rho_{J,j}]^\mathsf{T},
& \mathrm{EMA},\\
[P_N^{\rm kW},\,N_g,\,p^{\rm mm},\,\rho_{m,j}]^\mathsf{T},
& \mathrm{EMLA},
\end{cases}
\label{eq:lower_obj}
\end{equation}
where
\begin{equation}
\left\{
\begin{aligned}
\rho_{J,j}
&=
\frac{J_mN_g^2}
{J_{\mathrm{eq},j}^{\min,+}},\\
\rho_{m,j}
&=
\frac{J_m(2\pi N_g/p)^2}
{m_{\mathrm{eq},j}^{\min,+}} .
\end{aligned}
\right.
\label{eq:reflected_inertia}
\end{equation}
Here $J_m$ is motor inertia, while
$J_{\mathrm{eq},j}^{\min,+}$ and
$m_{\mathrm{eq},j}^{\min,+}$ are the minimum positive
equivalent rotational inertia and translational mass over the
task. The ratios discourage excessive reflected motor inertia.

Subject to the actuator and topology-dependent constraints, the
feasible discrete follower set is
\begin{equation}
\mathcal Y_{\ell,j}^{\rm feas}(\vect x,\vect m_a)=
\left\{
\vect y_j\in\mathcal Y_j(\vect x):
\vect G_{\ell,j}(\vect x,\vect m_a,\vect y_j)
\preceq\vect 0
\right\}.
\label{eq:lower_feasible_set}
\end{equation}
Rather than returning a single joint solution, each search
retains a bounded archive
$\mathcal A_j(\vect x,\vect m_a)
\subset\mathcal Y_{\ell,j}^{\rm feas}$.
Nondominated solutions are prioritized while additional feasible
alternatives are retained for system-level coordination.

\subsection{System-Level Package Coordination and Mass Feedback}

The joint archives are coordinated jointly rather than selecting
each actuator independently. Omitting their dependence on
$(\vect x,\vect m_a)$ for brevity, define
\begin{equation}
\left\{
\begin{aligned}
\mathcal A
&=\mathcal A_1\times\mathcal A_2\times\mathcal A_3,\\
\vect z(\vect y)
&=
[M_a(\vect y),\,C_a(\vect y),\,E_{\rm sup}(\vect y)]^\mathsf{T},
\qquad \vect y\in\mathcal A .
\end{aligned}
\right.
\label{eq:system_coordination}
\end{equation}
where each $\vect y$ is a realizable three-joint package and
$M_a$, $C_a$, and $E_{\rm sup}$ are its total actuator mass,
cost, and supply energy.

The three metrics are min--max normalized over $\mathcal A$,
and an equal-weight ideal-point compromise is selected:
\begin{equation}
\left\{
\begin{aligned}
\hat z_r(\vect y)&=
\frac{z_r(\vect y)-z_r^{\min}}
{z_r^{\max}-z_r^{\min}+\epsilon_z},
\qquad r=1,2,3,\\
\vect y_b^{(k)}&=
\arg\min_{\vect y\in\mathcal A}
\|\hat{\vect z}(\vect y)\|_2 .
\end{aligned}
\right.
\label{eq:package_selection}
\end{equation}
Thus, mass, cost, and energy enter the coordination with equal
explicit weighting after normalization.

For fixed $\vect x$, robot simulation, joint-level search, and
package coordination define the actuator-mass response
$\mathcal T_{\vect x}$. The coupling iteration is
\begin{equation}
\left\{
\begin{aligned}
\vect m_a^{(k+1)}
&=\mathcal T_{\vect x}(\vect m_a^{(k)})
=\vect m_a(\vect y_b^{(k)}),\\
e_m^{(k)}
&=
\frac{\|\vect m_a^{(k+1)}-\vect m_a^{(k)}\|_2}
{\|\vect m_a^{(k)}\|_2+\epsilon}.
\end{aligned}
\right.
\label{eq:mass_feedback}
\end{equation}
If $e_m^{(k)}>\epsilon_m$, the updated masses are returned to
the robot model and the actuator demands and package selection
are recomputed. Iteration terminates at
$e_m^{(k)}\leq\epsilon_m$, yielding the converged package
$\vect y_b^\star$ and mass vector
$\vect m_a^\star\approx
\mathcal T_{\vect x}(\vect m_a^\star)$.
The upper-level objectives and constraints are then evaluated
using these mass-consistent installed actuator masses.

\subsection{Upper-Level Objectives and Feasibility}

After actuator-mass convergence, the upper objectives are
\begin{equation}
\vect F_u(\vect x,\vect y_b^\star)=
\begin{bmatrix}
M_{\rm hw}\\
C_{\rm codesign}\\
E_{\rm sup}
\end{bmatrix}
=
\begin{bmatrix}
M_{\rm link}+M_a+M_{\rm wrist}\\
C_{\rm structure}+C_a\\
E_{\rm sup}
\end{bmatrix}.
\label{eq:upper_obj}
\end{equation}
The tool payload is excluded from $M_{\rm hw}$ but retained in
the dynamic and structural loading. Likewise, the fixed wrist
cost is omitted from $C_{\rm codesign}$ because it is common
to all designs.

A candidate is feasible only if the converged system satisfies
the prescribed $SE(3)$ task, kinematic and geometric
requirements, dynamic tracking and structural-stress limits,
actuator ratings, topology-dependent EMLA constraints, and
actuator-mass consistency. Constraint handling uses
feasible-first constrained Pareto dominance: feasible designs
are ranked by their objective vectors, whereas infeasible
designs are ordered by aggregate constraint violation. The upper search therefore returns a finite nondominated
approximation of mass-consistent structure--actuation designs,
rather than a scalar optimum. Both upper- and lower-level
multiobjective searches employ NSGA-II \cite{deb2002fast}.

\begin{algorithm}[t]
\caption{Multiobjective Bilevel Structure--Actuation Co-Design}
\label{alg:bilevel}
\footnotesize
\begin{algorithmic}[1]

\REQUIRE Mechanical bounds, actuator catalogue,
task data, $\vect m_a^{(0)}$, $K_{\max}$, $\epsilon_m$

\STATE Initialize a repaired upper-level population

\FOR{each upper NSGA-II generation}

    \FOR{each unevaluated mechanical design $\vect x$}

        \STATE Repair $\vect x$ and canonicalize inactive variables
        \STATE $\vect m_a \leftarrow \vect m_a^{(0)}$

        \FOR{$k=0,\ldots,K_{\max}-1$}

            \STATE Simulate robot using $(\vect x,\vect m_a)$
            and extract joint demand histories

            \FOR{$j=1,2,3$}
                \STATE Run topology-dependent constrained
                lower NSGA-II and construct $\mathcal A_j$
            \ENDFOR

            \STATE Form
            $\mathcal A_1\times\mathcal A_2\times\mathcal A_3$
            and select $\vect y_b^{(k)}$ using the normalized
            mass--cost--energy ideal point

            \STATE
            $\vect m_a^{(k+1)}
            \leftarrow \vect m_a(\vect y_b^{(k)})$

            \IF{$e_m^{(k)}\leq\epsilon_m$}
                \STATE \textbf{break}
            \ENDIF

            \STATE $\vect m_a\leftarrow\vect m_a^{(k+1)}$

        \ENDFOR

        \STATE Perform final package-mass-consistent evaluation
        \STATE Evaluate $\vect F_u$ and $\vect G_u$

    \ENDFOR

    \STATE Apply feasible-first nondominated sorting,
    crowding, selection, and variation

\ENDFOR

\RETURN Obtained feasible rank-zero designs

\end{algorithmic}
\end{algorithm}

\section{Results and Discussion}

\subsection{Numerical Setup}
The nominal study considers a \SI{100}{kg} tool payload and a
fixed \SI{50}{kg} spherical wrist, giving a distal carried mass
of \SI{150}{kg} in the dynamic and structural models.
Table~\ref{tab:setup} summarizes the mechanical search space
and numerical settings. The actuator catalogue contains 68
motor records, while gearbox ratios and EMLA screw leads are
discrete. Full-pose feasibility requires all 12 prescribed
$SE(3)$ targets, a minimum joint-limit and wrist-singularity
margin of $5^{\circ}$, and a main-arm condition number not
exceeding 100. The dynamic and structural limits are
$e_{\max}\leq\SI{20}{mm}$ and
$\sigma_{\max}\leq\SI{250}{MPa}$, respectively.

For reporting, the co-design cost is normalized as
\begin{equation}
C_{\rm idx}=
\frac{C_{\rm codesign}}{C_{\rm ref}},
\label{eq:cost_index}
\end{equation}
where $C_{\rm ref}$ is the lowest co-design cost observed
across the nominal and payload-screening datasets. The same
reference is used for all reported cost indices. The nominal
campaign uses the full simulation resolution
($\Delta t$ scale 1.0) and fixed upper- and lower-level random
seeds for reproducibility.

\begin{table}[t]
\caption{Principal Design and Optimization Settings}
\label{tab:setup}
\centering
\footnotesize
\setlength{\tabcolsep}{3.0pt}
\begin{tabular}{@{}p{0.43\columnwidth}p{0.52\columnwidth}@{}}
\toprule
Item & Value \\
\midrule
$L_1$; $L_2,L_3$
& $[0.8,2.0]$ m; $[2.5,5.0]$ m \\

$\nu_2,\nu_3$
& rotary / 3-bar / 4-bar \\

$A_2,A_3$; $B_2,B_3$
& $[0.60,0.90]$ m; $[0.30,0.90]$ m \\

$\alpha_2$; $\alpha_3$
& $[0.01,0.50]$ rad; $[0.01,0.20]$ rad \\

Motor / $N$ / $p$
& 68 motors; $N,p\in\{1,\ldots,30\}$,
$p$ in mm/rev \\

Tool / fixed wrist
& 100 kg / 50 kg \\

Upper NSGA-II
& population 20; generation cap 15; seed 101 \\

Lower NSGA-II
& population 18; generation cap 12; seed 1001 \\

Joint archive
& at most 12 retained alternatives per joint \\

Mass coupling
& $\vect m_a^{(0)}=[15,15,15]^\mathsf{T}$ kg;
$K_{\max}=6$; $\epsilon_m=5\times10^{-3}$ \\

Simulation
& full $\Delta t$ scale 1.0;
$e_{\max}\leq20$ mm;
$\sigma_{\max}\leq250$ MPa \\
\bottomrule
\end{tabular}
\end{table}

\subsection{Nominal 100-kg Tool-Payload Co-Design}

\begin{figure*}[t]
    \centering
    \includegraphics[width=0.85\textwidth]
    {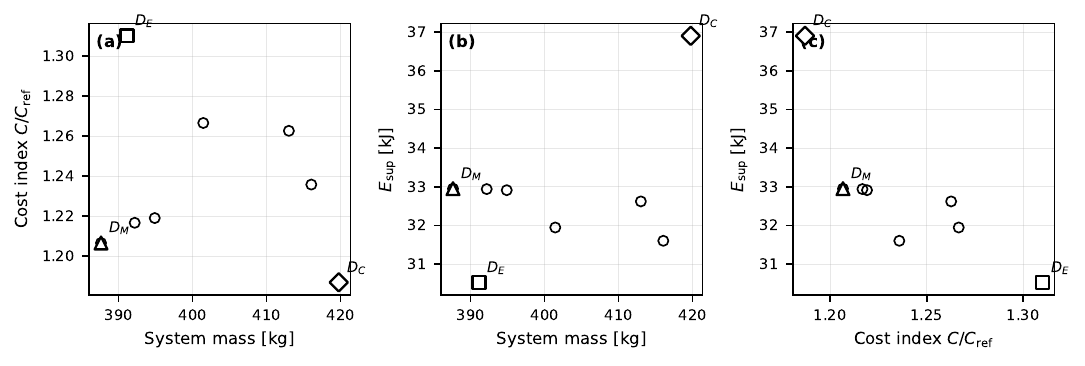}
    \caption{Nominal 100-kg nondominated set:
    (a) hardware mass versus normalized co-design cost,
    (b) hardware mass versus supply energy, and
    (c) normalized cost versus supply energy.
    $D_M$, $D_C$, and $D_E$ denote the minimum-mass,
    minimum-cost, and minimum-energy designs.}
    \label{fig:nominal_tradeoffs}
\end{figure*}

The fixed-budget nominal search evaluated 320 unique
upper-level designs and returned eight feasible nondominated
solutions satisfying all kinematic, dynamic, structural,
actuator, and mass-consistency requirements.
Figure~\ref{fig:nominal_tradeoffs} shows their three-objective
trade-offs.

The set spans $387.62$--$419.74$~kg in hardware mass,
$1.187$--$1.310$ in normalized co-design cost, and
$30.53$--$36.90$~kJ in supply energy. The extrema are
\[
\begin{array}{c|ccc}
 & M_{\rm hw}\,[\mathrm{kg}] & C_{\rm idx} & E_{\rm sup}\,[\mathrm{kJ}]\\
\hline
D_M & 387.62 & 1.207 & 32.95\\
D_C & 419.74 & 1.187 & 36.90\\
D_E & 391.13 & 1.310 & 30.53
\end{array}
\]
showing that improvement of one objective generally requires
sacrifice in another.

\begin{table*}[t]
\caption{Representative Nominal Co-Designs}
\label{tab:representative_designs}
\centering
\scriptsize
\setlength{\tabcolsep}{3.2pt}
\begin{tabular}{@{}lccccccc@{}}
\toprule
Design &
$J_2/J_3$ &
$[L_1,L_2,L_3]$ [m] &
$[A_2,B_2,\alpha_2]$ &
$[A_3,B_3,\alpha_3]$ &
$J_1$: M/$P_N$/$N_g$ &
$J_2$: M/$P_N$/$N_g$/$p$ &
$J_3$: M/$P_N$/$N_g$/$p$ \\
\midrule
$D_M$ &
4-bar/3-bar &
$[1.145,4.961,2.543]$ &
$[0.772,0.370,0.329]$ &
$[0.817,0.678,0.100]$ &
M01/11/15 &
M02/4/6/7 &
M03/2.2/3/5 \\

$D_E$ &
4-bar/3-bar &
$[1.145,4.961,2.543]$ &
$[0.683,0.828,0.060]$ &
$[0.817,0.699,0.100]$ &
M04/3/26 &
M01/11/6/7 &
M03/2.2/3/5 \\

$D_C$ &
4-bar/4-bar &
$[1.213,4.961,2.682]$ &
$[0.683,0.828,0.100]$ &
$[0.654,0.397,0.100]$ &
M05/5.5/15 &
M06/2.2/9/13 &
M07/2.2/6/9 \\
\bottomrule
\end{tabular}

\vspace{0.5mm}
\parbox{0.98\textwidth}{\scriptsize
$A_j,B_j$ are in m and $\alpha_j$ in rad.
M01--M07 denote anonymized catalogue motors;
$P_N$ is in kW, $N_g$ is the gearbox ratio, and $p$ is the
EMLA screw lead in mm/rev.}
\end{table*}

Table~\ref{tab:representative_designs} exposes the
structure--actuation coupling behind these objective extremes.
Although $D_M$ and $D_E$ share the same 4-bar/3-bar topology
and principal link dimensions, their linkage geometry and
drive packages differ markedly: $D_M$ uses an 11-kW,
$N_g=15$ motor at $J_1$ and a 4-kW motor at $J_2$, whereas
$D_E$ uses a 3-kW, $N_g=26$ motor at $J_1$ and an 11-kW
motor at $J_2$. Their $J_3$ selections remain nearly
unchanged. In contrast, $D_C$ adopts a 4-bar/4-bar topology
and 2.2-kW motors at both linearly actuated joints. Thus,
similar arm dimensions can yield substantially different
electromechanical realizations, confirming that the trade-off
is governed jointly by mechanical architecture and discrete
actuator selection.

\subsection{Payload Sensitivity}

Payload sensitivity was screened at tool loads of
20, 60, 100, and 140~kg using a common reduced fixed search
budget and the fixed 50-kg wrist. These results therefore
indicate payload-dependent trends rather than converged
payload-specific optima.

\begin{figure*}[t]
    \centering
    \includegraphics[width=0.85\textwidth]
    {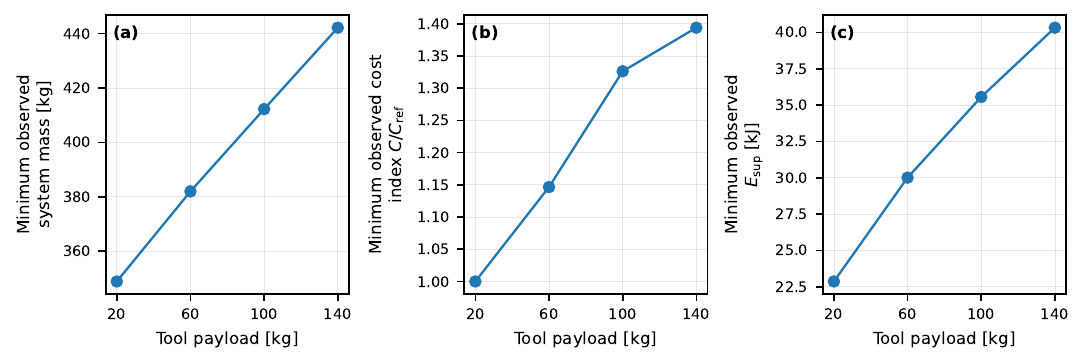}
    \caption{Payload screening: minimum observed
    (a) hardware mass, (b) normalized co-design cost, and
    (c) supply energy within each obtained nondominated set.
    All cases use the same reduced fixed search budget.}
    \label{fig:payload_sensitivity}
\end{figure*}

As the tool payload increases from 20 to 140~kg, the minimum
observed hardware mass rises from 348.87 to 442.16~kg, the
cost index from 1.000 to 1.394, and supply energy from
22.86 to 40.33~kJ. The competitive mechanism topology also
changes with load: the 20-kg set contains 3-bar/4-bar and
4-bar/4-bar designs; the 60- and 100-kg sets contain
3-bar/4-bar, 4-bar/3-bar, and 4-bar/4-bar designs; and the
140-kg set retains only 3-bar/4-bar and 4-bar/3-bar designs.
Moreover, the minimum-mass, minimum-cost, and minimum-energy
solutions need not share the same topology, showing that
payload influences both actuator demand and the preferred
mechanical architecture.

\subsection{Effect of Actuator-Mass Feedback}

Actuator-mass coupling was isolated through a fixed-design
ablation of $D_M$, $D_E$, and $D_C$. For each design,
actuators were sized once from the common initial estimate
$\vect m_a^{(0)}=[15,15,15]^\mathsf{T}$~kg; the selected
package was then frozen and re-evaluated using its installed
masses without further actuator reselection. This one-pass
baseline was compared with the fully coupled solution using
\begin{equation}
\begin{aligned}
\Delta z &=
100\frac{z_{\rm 1pass}-z_{\rm coupled}}{z_{\rm coupled}},\\
\epsilon_E &=
100\frac{E_{\rm sup,installed}-E_{\rm sup,pred}}
{E_{\rm sup,pred}} .
\end{aligned}
\label{eq:ablation_metrics}
\end{equation}

\begin{table}[t]
\caption{Fixed-Design Actuator-Mass-Feedback Ablation}
\label{tab:mass_feedback_ablation}
\centering
\scriptsize
\setlength{\tabcolsep}{3.2pt}
\begin{tabular}{@{}lccccc@{}}
\toprule
Design &
$e_m^{(0)}$ &
$\Delta M$ [\%] &
$\Delta C$ [\%] &
$\Delta E_{\rm sup}$ [\%] &
$\epsilon_E$ [\%] \\
\midrule
$D_M$ & 2.655 & +0.47 & -0.71 & +9.40 & +1.89 \\
$D_E$ & 2.836 & +2.13 & -3.81 & +8.09 & +1.75 \\
$D_C$ & 3.367 & +0.43 & +10.20 & -8.80 & +3.67 \\
\bottomrule
\end{tabular}
\end{table}

The first-update discrepancies,
$e_m^{(0)}=2.655$--$3.367$, quantify differences between
the initial estimates and selected package masses.
All frozen one-pass packages satisfy the physical constraints
after re-evaluation with their installed masses.
Actuator reselection changes the realized trade-off:
hardware mass changes by $0.43$--$2.13\%$, co-design cost by
$-3.81$ to $+10.20\%$, and supply energy by $-8.80$ to
$+9.40\%$. The discrete follower response changes as well;
for example, $D_M$ changes its $J_2$ screw lead from
10 to 7~mm/rev after coupling, while $D_E$ changes both its
$J_1$ package and $J_2$ screw selection. Thus, actuator-mass feedback makes actuator selection
consistent with the demand histories induced by the selected
packages, although it does not systematically improve every
individual objective.

For $D_M$, the mass iteration evolves from
$[15,15,15]^\mathsf{T}$ to
$[77.36,43.72,21.49]^\mathsf{T}$ and then
$[77.86,41.40,21.49]^\mathsf{T}$~kg, which is reproduced on
the following iteration. The remaining reported nominal
designs likewise satisfy the mass-convergence criterion in
three iterations.

\subsection{Constraint and Physical Validation}

All eight nominal nondominated designs satisfy the prescribed
physical constraints and all 12 full-pose targets. Across the
set, the worst-case tracking error and structural stress are
9.73~mm and 128.35~MPa, below their respective limits of
20~mm and 250~MPa. The minimum joint-limit and EMLA stroke
margins are $48.05^{\circ}$ and 34.41~mm, while the maximum
main-arm condition number is 6.11. Maximum torque, speed, and
power utilizations are 0.995, 0.983, and 0.946, respectively.
Actuator-mass coupling converges for every reported design in
three iterations with zero final residual and no detected cycle.

Because the evolutionary search uses a prescribed finite
budget, the reported designs constitute a nondominated
approximation rather than a globally converged Pareto front.

\section{Conclusion}

A multiobjective bilevel structure--actuation co-design
framework was developed for all-electric heavy-duty robotic
manipulators. Mixed mechanism topology and continuous geometry
are optimized jointly with discrete actuator packages, while
system-level package coordination and actuator-mass feedback
ensure consistency between actuator sizing and robot dynamics.

For the nominal 100-kg tool-payload case, the fixed-budget
search produced eight feasible nondominated designs satisfying
all 12 prescribed $SE(3)$ poses, with maximum structural stress of 128.35~MPa. Actuator-mass
coupling converged for every reported design in three
iterations. The fixed-design ablation showed that one-pass
sequential sizing changes hardware mass by up to 2.13\%,
co-design cost by 10.20\%, and supply energy by 9.40\% in
magnitude, demonstrating that actuator-mass feedback can alter
both the selected discrete package and the resulting system
trade-off. Payload screening further showed increasing
mass--cost--energy demand and changes in competitive mechanism
topology with carried load.

The reported solutions constitute a fixed-budget nondominated
approximation rather than a globally converged Pareto front.
Current limitations include the engineering structural and
motor-loss models, the fixed-design wrist, and omission of
wrist-actuator energy. Future work will address higher-fidelity
component and thermal models, robust co-design under
uncertainty, and experimental validation.

\bibliographystyle{IEEEtran}
\bibliography{references}

\end{document}